\documentclass[11pt,a4paper]{article}
\usepackage{times,latexsym}
\usepackage[T1]{fontenc}
\usepackage[a4paper]{geometry}
\usepackage{acl2020}
\usepackage{times}
\usepackage{latexsym}
\usepackage{amsmath}
\usepackage{mathtools}
\usepackage{amssymb}
\usepackage{multirow}
\usepackage{url}
\usepackage{booktabs}
\usepackage{etoolbox}
\usepackage[]{hyperref}
\usepackage{comment}
\usepackage[compact]{titlesec}
\usepackage{graphicx}

\makeatletter
\newcommand{\@emptybiblabel}[1]{}
\makeatother

\aclfinalcopy 
\def\aclpaperid{3486} 

\title{Predictive Biases in Natural Language Processing Models:\\ A Conceptual Framework and Overview}

\author{Deven Shah \hspace{3em} H. Andrew Schwartz\\ 
        Stony Brook University\hspace{2em} \\
        \texttt{\{dsshah,has\}@cs.stonybrook.edu}\hspace{2em}
        \And
        Dirk Hovy\\
        Bocconi University\\
        \texttt{dirk.hovy@unibocconi.it}
        }
\date{}

\begin{document}
\maketitle

\begin{abstract}
An increasing number of natural language processing papers address the effect of bias on predictions, introducing mitigation techniques at different parts of the standard NLP pipeline (data and models).
However, these works have been conducted individually, without a unifying framework to organize efforts within the field. 
This situation leads to repetitive approaches, and focuses overly on bias \textit{symptoms/effects}, rather than on their \textit{origins}, which could limit the development of effective countermeasures.
In this paper, we propose a unifying \textit{predictive bias framework for NLP}. 
We summarize the NLP literature and suggest general mathematical definitions of predictive bias. We differentiate two consequences of bias: \textit{outcome disparities} and \textit{error disparities}, as well as four potential origins of biases: \textit{label bias}, \textit{selection bias}, \textit{model overamplification}, and \textit{semantic bias}.
Our framework serves as an overview of predictive bias in NLP, integrating existing work into a single structure, and providing a conceptual baseline for improved frameworks. 
\end{abstract}

\section{Introduction}

Predictive models in NLP are sensitive to a variety of (often unintended) biases throughout the development process. As a result, fitted models do not generalize well, incurring performance and reliability losses on unseen data. They also have socially undesirable effects by systematically under-serving or mispredicting certain user groups. 

The general phenomenon of biased predictive models in NLP is not recent. The community has long worked on the domain adaptation problem~\cite{jiang2007instance,daume2007frustratingly}: models fit on newswire data do not perform well on social media and other text types. This problem arises from the tendency of statistical models to pick up on non-generalizable signals during the training process. 
In the case of domains, these non-generalizations are words, phrases, or senses that occur in one text type, but not another.

However, this kind of variation is not just restricted to text domains: it is a fundamental property of \textit{human-generated} language: we talk differently than our parents or people from a different part of our country, etc.~\cite{pennebaker2003words,eisenstein2010latent,kern2016gaining}. In other words, language reflects the diverse demographics, backgrounds, and personalities of the people who use it. 
While these differences are often subtle, they are distinct and cumulative~\cite{trudgill2000sociolinguistics,kern2016gaining,pennebaker2011secret}. 
Similar to text domains, this variation can lead models to pick up on patterns that do not generalize to other author-demographics, or to rely on undesirable word-demographic relationships.

Bias may be an inherent property of any NLP system (and broadly any statistical model), but this is not per se negative. In essence, biases are priors that inform our decisions (a dialogue system designed for elders might work differently than one for teenagers). Still, undetected and unaddressed, biases can lead to negative consequences:
There are aggregate effects for demographic groups, which combine to produce \textit{predictive bias}. I.e., the label distribution of a predictive model reflects a human attribute in a way that diverges from a theoretically defined ``ideal distribution.'' For example, a Part Of Speech (POS) tagger reflecting how an older generation uses words \cite{hovy2015tagging} diverges from the population as a whole.

\begin{figure*}[tbh!]
    \centering
    \includegraphics[width=\textwidth]{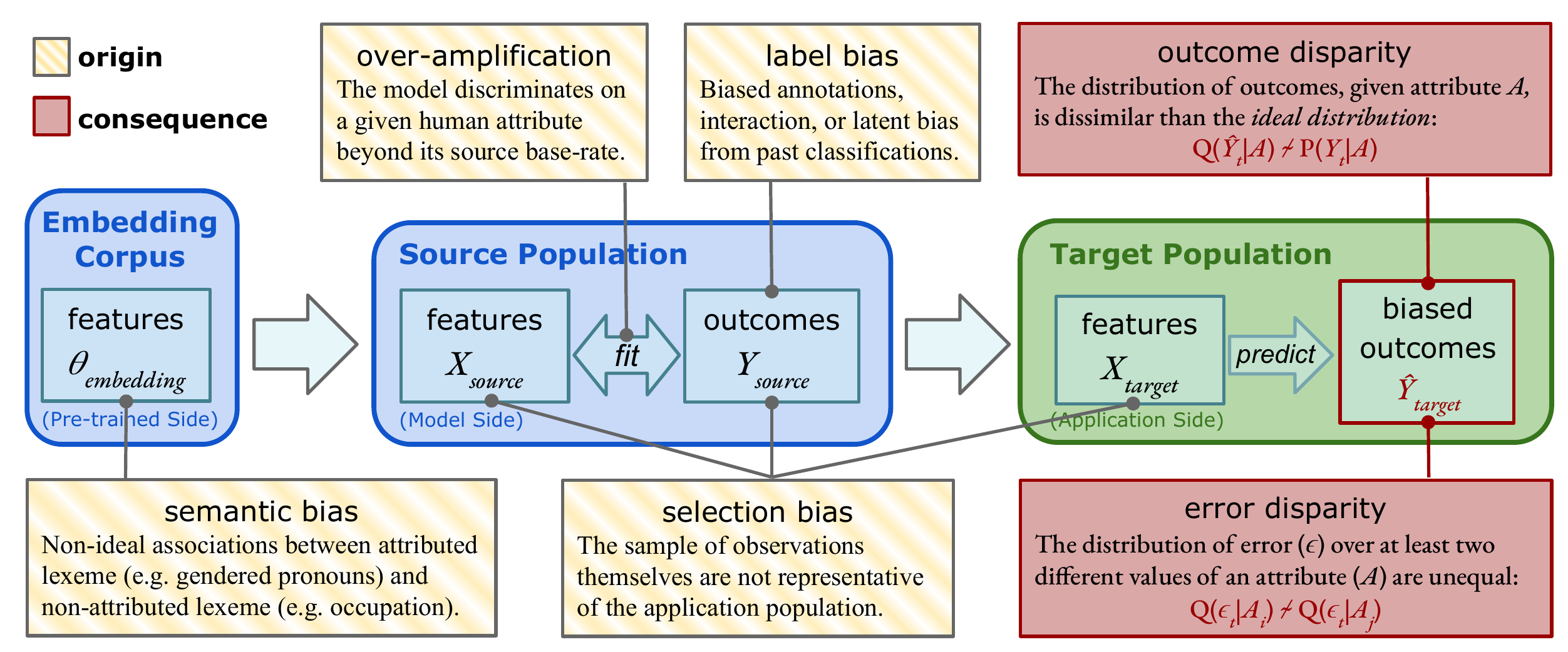}
    \caption{\textit{The Predictive Bias Framework for NLP}: Depiction of where bias may originate within a standard supervised NLP pipeline. Evidence of bias is seen in $\hat{y}$ via \textit{outcome disparity} and \textit{error disparity}.}
    \label{fig:origins}
\end{figure*}

A variety of papers have begun to address countermeasures for predictive biases ~\cite{li2018towards,elazar2018adversarial,coavoux2018privacy}.\footnote{An even more extensive body of work on \textit{fairness} exists as part of the FAT* conferences, which goes beyond the scope of this \textit{biased-focused} paper. Note also that while bias is an ethical issue and contributes to many papers in the ethics in NLP area, the two should not be conflated: ethics covers more than bias.} 
Each identifies a specific bias and countermeasure on their terms, \textbf{but} it is often not explicitly clear which bias is addressed, where it originates, or how it generalizes. 
There are multiple sources from which bias can arise within the predictive pipeline, and methods proposed for one specific bias often do not apply to another. 
As a consequence, much work has focused on bias \textit{effects and symptoms} rather than their \textit{origins}.
While it is essential to address the effects of bias, it can leave the fundamental origin unchanged~\cite{gonen-goldberg-2019-lipstick-pig}, requiring researchers to rediscover the issue over and over.
The ``bias'' discussed in one paper may, therefore, be quite different than that in another.\footnote{Quantitative social science offers a background for  bias~\cite{berk1983introduction}. However, NLP differs fundamentally in analytic goals (namely out-of-sample prediction for NLP versus parameter inference for hypothesis testing in social science) that bring about NLP-specific situations: biases in word embeddings, annotator labels, or predicting over-amplified demographics.}  

A shared definition and framework of predictive bias can unify these efforts, provide a common terminology, help to identify underlying causes, and allow coordination of countermeasures~\cite{sun-etal-2019-mitigating}. 
However, such a general framework had yet to be proposed within the NLP community.

To address these problems, we suggest a joint conceptual framework, depicted in Figure \ref{fig:origins}, outlining and relating the different origins of bias. We base our framework on an extensive survey of the relevant NLP literature, informed by selected works in social science and adjacent fields.
We identify four distinct sources of bias: \textbf{selection bias}, \textbf{label bias}, \textbf{model overamplification}, and \textbf{semantic bias}. We can express all of these as differences between (a) a ``true'' or intended distribution (e.g., over users, labels, or outcomes), and (b) the distribution used or produced by the model. These cases arise at specific points within a typical predictive pipeline: embeddings, source data, labels (human annotators), models, and target data. 
We provide quantitative definitions of predictive bias in this framework intended to make it easier to: (a) identify biases (because they can be classified), (b) develop countermeasures (because the underlying problem is known), and (c) compare biases and countermeasures across papers. 
We hope this paper will help researchers spot, compare, and address bias in all its various forms.

\paragraph{Contributions}
Our primary contributions include: (1) a conceptual framework for identifying and quantifying predictive bias and its origins within a standard NLP pipeline,
(2) a survey of biases identified in NLP models,  and (3) a survey of methods for countering bias in NLP organized within our conceptual framework.

\section{Definition - Two Types of Disparities}
Our definition of \textit{predictive bias} in NLP builds on its definition within the literature on standardized testing (i.e., SAT, GRE, etc.) Specifically, 
\newcite{swinton1981predictive} states: 
\begin{quote}
\begin{small}
\textit{By ``predictive bias," we refer to a situation in which a [predictive model] is used to predict a specific criterion for a particular population, and is found to give systematically different predictions for subgroups of this population who are in fact identical on that specific criterion.}\footnote{We have substituted ``test" with ``predictive model''.}
\end{small}
\end{quote}

We generalize Swinton's definition in two ways: 
First, to align notation with standard supervised modeling, we say there are both $Y$ (a random variable representing the ``true'' values of an outcome) and  $\hat{Y}$ (a random variable representing the predictions. 
Next, we allow the concept to apply to differences associated with continuously-valued human attributes rather than simply discrete subgroups of people.\footnote{``Attributes'' include both continuously valued user-level variables, like age, personality on a 7-point scale, etc. (also referred to as ``dimensional'' or ``factors''), and discrete categories like membership in an ethnic group. Psychological research suggests that people are better represented by continuously valued scores, where possible, than discrete categories~\cite{baumeister2007psychology,widiger2005diagnostic,mccrae1989reinterpreting}. In NLP, \newcite{lynn2017human} shows benefits from treating user-level attributes as continuously when integrating into NLP models.}
Below, we define two types of measurable systematic differences (i.e. ``disparities''): (1) a systematic difference between $Y$ and $\hat{Y}$ ( \textit{outcome disparity}) and (2) a difference in error ($\epsilon = |Y - \hat{Y}$) \textit{error disparity}, both as a function of a given human attribute, $A$.

\paragraph{Outcome disparity.} Formally, we say an \textit{outcome disparity} exists for outcome, $Y$, a domain $D$ (with values $source$ or $target$), and with respect to attribute, $A$, when the distribution of the predicted outcome ($Q(\hat{Y}_D|A_D)$) is dissimilar to a given theoretical \textit{ideal distribution} ($P(Y_D|A_D)$):  
$$Q(\hat{Y}_D|A_D) \nsim P(Y_D|A_D)$$

The \textit{ideal distribution} is specific to the target application. Our framework allows researchers to use their own criteria to determine this distribution. However, the task of doing so may be non-trivial. 
First, the current distribution within a population may not be accessible. Even when it is, it may not be what most consider the ideal distribution (e.g., the distribution of gender in computer science and the associated disparity of NLP models attributing male pronouns to computer scientists more frequently~\cite{hovy2015demographic}). 
Second, it may be difficult to come to an agreed-upon ideal distribution from a moral or ethical perspective. 
In such a case, it may be helpful to use an ideal ``direction,'' rather than specifying a specific distribution (e.g., moving toward a uniform distribution of pronouns associated with computer science).  
Our framework should enable its users to apply evolving standards and norms across NLP's many application contexts.

A \textbf{prototypical example of \textit{outcome disparity}} is gender disparity in image captions. \newcite{zhao2017men} and \newcite{hendricks2018women} demonstrate a systematic difference with respect to gender in the outcome of the model, $\hat{Y}$ even when taking the source distribution as an ideal target distribution: $Q(\hat{Y}_{target}|gender) \nsim Q(Y_{target}|gender) \sim  Q(Y_{source}|gender)$. As a result, captions over-predict females in images with ovens and males in images with snowboards. 

\paragraph{Error disparity.} We say there is an \textit{error disparity} when model predictions have larger error for individuals with a given user attribute (or range of attributes in the case of continuously-valued attributes).
Formally, the error of a predicted distribution is
$$\epsilon_D = |Y_D - \hat{Y}_D|$$
If there is a difference in $\epsilon_D$ over at least two different values of an attribute $A$ (assuming they have been adequately sampled to establish a distribution of $\epsilon_D$) then there is an error disparity.

$$Q(\epsilon_D|A_i) \nsim Q(\epsilon_D|A_j)$$

In other words, the error for one group might systematically differ from the error for another group, e.g., the error for green people differs from the error for blue people. Under unbiased conditions, the difference would be equal. This formulation allows us to capture both the discrete case (arguably more common in NLP, for example, in POS tagging) and the continuous case (for example, in age or income prediction).
\\\\
We propose that if either of these two disparities exist in our target application, then there is a \textit{predictive bias}.
Note that predictive bias is then a property of a model \textit{given} a specific application, rather than merely an intrinsic property of the model by itself. 
This definition mirrors predictive bias in standardized testing \cite{swinton1981predictive}: ``a [predictive model] cannot be called biased without reference to a specific prediction situation; thus, the same instrument may be biased in one application, but unbiased in another."

A \textbf{prototypical example of \textit{error disparity}} is the ``Wall Street Journal Effect'' -- a systematic difference in error as a function of demographics, first documented by \newcite{hovy2015tagging}. In theory, POS tagging errors increase the further an author's demographic attributes differ from the average WSJ author of the 1980s and 1990s (on whom many POS taggers were trained -- a selection bias, discussed next). 
Work by  \newcite{sap-etal-2019-risk} shows error disparity from a different origin, namely unfairness in hate speech detection. They find that annotators for hate speech on social media make more mistakes on posts of black individuals. Contrary to the case above, the disparity is not necessarily due to a difference between author and annotator population (a selection bias). Instead, the label disparity stems from annotators failing to account for the authors' racial background and sociolinguistic norms.

\paragraph{Source and Target Populations.}
An important assumption of our framework is that disparities are dependent on the population for which the model will be applied. 
This assumption is reflected in distinguishing a separate ``target population'' from the ``source population'' on which the model was trained. 
In cross-validation over random folds, models are trained and tested over the same population. However, in practice, models are often applied to novel data that may originate from a different population of people. 
In other words, the disparity may exist as a model property for one application, but not for another. 

\paragraph*{Quantifying disparity.}
Given the definitions of the two types of disparities, we can quantify bias with well-established measures of distributional divergence or deviance. Specifically, we suggest the Log-likelihood ratio as a central metric:
$$D(Y, \hat{Y}| A) = 2(log(p(Y|A)) - log(p(\hat{Y}|A))) $$
where $p(Y|A)$ is the specified ideal distribution (either derived empirically or theoretically) and $p(\hat{Y}|A)$ is the distribution within the data. For error disparity the ideal distribution is always the \textit{Uniform} and $\hat{Y}$ is replaced with the error.  KL divergence ($D_{KL}[P(\hat{Y}|A) P(Y|A)]$) can be used as a secondary, more scalable alternative. 

Our measure above attempts to synthesize metrics others have used in works focused on specific biases. 
For example, the definition of outcome disparity is analogous to that used for semantic bias.  \newcite{kurita2019measuring} quantify bias in embeddings as the difference in log probability score when replacing words suspected to carry semantic differences (`he', `she') with a mask: 

\begin{small}
\begin{equation}
\begin{aligned}
\log(P([Mask] = ``\langle PRON\rangle " | 
[Mask]\ is\ ``\langle NOUN\rangle"))\ -\\
\log(P([Mask] = ``\langle PRON\rangle" | 
[Mask]\ is\ [Mask]))) \nonumber
\end{aligned}
\end{equation}
\end{small}
$\langle NOUN\rangle$ is replaced with a specific noun to check for semantic bias (e.g., an occupation), and $\langle PRON\rangle$ is an associated demographic word (e.g., ``he'' or ``she'').

\section{Four Origins of Bias}
But what leads to an outcome disparity or error disparity? We identify four points within the standard supervised NLP pipeline where bias may originate: (1) the training labels (\textit{label bias}), (2) the samples used as observations --- for training or testing (\textit{selection bias}), (3) the representation of data (\textit{semantic bias}), or (4) due to the fit method itself (\textit{overamplification}).

\paragraph{Label Bias}
Label bias emerges when the distribution of the dependent variable in the data source diverges substantially from the ideal distribution: 
$$Q(Y_s|A_s) \nsim P(Y_s|A_s)$$
Here, the labels themselves are erroneous concerning the demographic attribute of interest (as compared to the source distribution). 
Sometimes, this bias is due to a non-representative group of annotators~\cite{joseph2017constance}. In other cases, it may be due to a lack of domain expertise \cite{plank2014learning}, or due to preconceived notions and stereotypes held by the annotators~\cite{sap-etal-2019-risk}.

\paragraph{Selection bias.} Selection bias emerges due to non-representative observations. I.e., when the users generating the training (source) observations differ from the user distribution of the target, where the model will be applied. Selection bias (sometimes also referred to as \textit{sample bias}) has long been a concern in the social sciences. At this point, testing for such a bias is a fundamental consideration in study design~\cite{berk1983introduction,culotta2014reducing}. Non-representative data is the origin for selection bias. 

Within NLP, some of the first works to note demographic biases were due to a selection bias~\cite{hovy2015tagging,jorgensen2015challenges}. A prominent example is the so-called ``Wall Street Journal effect'', where syntactic parsers and part-of-speech taggers are most accurate over language written by middle-aged white men. The effect occurs because this group happened to be the predominant authors' demographics of the WSJ articles, which are traditionally used to train syntactic models~\cite{garimella-etal-2019-womens}. The same effect was reported for language identification difficulties for African-American Vernacular English \cite{blodgett2017racial,jurgens2017incorporating}.

The predicted output is dissimilar from the ideal distribution, leading, for example, to lower accuracy for a given demographic, since the source did not reflect the \textit{ideal distribution}. 
We say that the distribution of human attribute, $A$, within the source data, $s$, is dissimilar to the distribution of $A$ within the target data, $t$: 
$$Q(A_s) \nsim P(A_t)$$ 

Selection bias has several peculiarities. 
First, it is dependent on the \textit{ideal distribution} of the target population, so a model may have selection bias for one application (and its associated target population), but not for another. 
Also, consider that either the source features ($X_s$) or source labels ($Y_s$) may be \textbf{non-representative}. In many situations, the distributions for the features and labels are the same. However, there are some cases where they diverge. For example, when using features from age-biased tweets, but labels from non-biased census surveys.
In such cases, we need to take multiple analysis levels into account: corrections can be applied to user features as they are aggregated to communities \cite{almodaresi2017distribution}. 
The consequences could be both \textit{outcome} and \textit{error disparity}.

One of the challenges in addressing selection bias is that we can not know \textit{a priori} what sort of (demographic) attribute will be important to control. Age and gender are well-studied, but others might be less obvious. We might someday realize that a formerly innocuous attribute (say, handedness) turns out to be relevant for selection biases. This problem is known as The Known and Unknown Unknowns.
\begin{quote}
\textit{As we know, there are known knowns: there are things we know we know. We also know there are known unknowns: that is to say, we know there are some things we do not know. But there are also unknown unknowns: the ones we don't know we don't know.}\\
--- Donald Rumsfeld
\end{quote}
We will see later how better documentation can help future researchers address this problem.

\begin{table}[]
    \centering
    \begin{tabular}{lp{0.16\columnwidth}|p{0.33\columnwidth}|p{0.22\columnwidth}}
         &      &  \multicolumn{2}{c}{\sc{Annotation}}\\
         &      & \textbf{incorrect} & \textbf{correct}  \\
    \midrule
         \multirow{2}{*}{\rotatebox[origin=c]{90}{\sc{Sample\hspace{4pt}}}} & \textbf{not-repr.} & selection bias, label bias &    selection bias\\
         \midrule
         
         & \textbf{repr.} & label bias & no bias   
    \end{tabular}
    \caption{Interaction between selection and label bias under different conditions for sample representativeness and annotation quality}
    \label{tab:bias_matrix}
\end{table}

\paragraph{Overamplification.}
Another source of bias can occur even when there is no label or selection bias. In \textit{overamplification}, a model relies on a small difference between human attributes with respect to the objective (even an acceptable difference matching the ideal distribution), but amplifies this difference to be much more pronounced in the predicted outcomes.
The origins of overamplification are during learning itself. The model learns to pick up on imperfect evidence for the outcome, which brings out the bias. 

Formally, in \textit{overamplification} the predicted distribution ($Q(\hat{Y}_s|A_s)$) is dissimilar to the source training distribution ($Q(Y_s|A_s)$) with respect to a human attribute, $A$. The predicted distribution is therefore also dissimilar to the target \textit{ideal distribution}:
$$Q(\hat{Y}_s|A_s) \nsim Q(Y_s|A_s) \sim P(Y_t|A_t)$$

For example, \newcite{yatskar2016situation} found that in the imSitu image captioning data set, 58\% of captions involving a person in a kitchen mention women. However, standard models trained on such data end up predicting people depicted in kitchens as women 63\% of the time~\cite{zhao2017men}. 
In other words, an error in generating a gender reference within the text (e.g., ``\textit{A} \textbf{[woman $\|$ man]} \textit{standing next to a counter-top}'') males an incorrect female reference much more common.

The occurrence of overamplification in the absence of other biases is an important motivation for countermeasures. It does not require bias on the part of the annotator, data collector, or even the programmer/data analyst (though it can escalate existing biases and the models' statistical discrimination along a demographic dimension). In particular, it extends countermeasures beyond the point some authors have made, that they are merely cosmetic and do not address the underlying cause: biased language in society~\cite{gonen-goldberg-2019-lipstick-pig}.

\paragraph{Semantic bias.} 
Embeddings (i.e., vectors representing the meaning of words or phrases) have become a mainstay of modern NLP, providing more flexible representations that feed both traditional and deep learning models.
However, these representations often contain unintended or undesirable associations and societal stereotypes (e.g., connecting medical doctors more frequently to male pronouns than female pronouns, see \newcite{bolukbasi2016man,caliskan2017semantics}). 
We adopt the term used for this phenomenon by others, ``semantic bias''.

Formally, we attribute semantic bias to the parameters of the embedding model ($\theta_{emb}$).
Semantic bias is a unique case since it indirectly affects both \textit{outcome disparity} and \textit{error disparity} by creating other biases, such as overamplification~\cite{yatskar2016situation,zhao2017men} or diverging words associations within embeddings or language models~\cite{bolukbasi2016man,rudinger2018gender}. 
However, we distinguish it from the other biases, since the population does not have to be people, but rather words in contexts that yield non-ideal associations. 
For example, the issue is not (only) that a particular gender authors more of the training data for the embeddings. Instead, that gendered pronouns are mentioned alongside occupations according to a non-ideal distribution (e.g., texts talk more about male doctors and female nurses than vice versa).
Furthermore, pre-trained embeddings are often used without access to the original data (or the resources to process it). We thus suggest that embedding models themselves are a distinct source of bias within NLP predictive pipelines.

They have consequently received increased attention, with dedicated sessions at NAACL and ACL 2019.
As an example, \newcite{kurita2019measuring} quantify human-like bias in BERT. Using the Gender Pronoun Resolution (GPR) task, they find that, even after balancing the data set, the model predicts no female pronouns with high probability. 
Semantic bias is also of broad interest to the social sciences as a diagnostic tool (see Section \ref{sec:appendix}). However, their inclusion in our framework is not for reasons of social scientific diagnostics, but rather to guide mindful researchers where to look for problems.

\paragraph{Multiple Biases.}
Biases occur not only in isolation, but they also compound to increase their effects. Label and selection bias can -- and often \textit{do} -- interact, so it can be challenging to distinguish them. Table \ref{tab:bias_matrix} shows the different conditions to understand the boundaries of one or another. 

Consider the case where a researcher chooses to balance a sentiment data set for a user attribute, e.g., age. This decision can directly impact the label distribution of the target variable. E.g., because the positive label is over-represented in a minority age group. Models learn to exploit this confounding correlation between age and label prevalence and magnify it even more. The resulting model may be useless, as it only captures the distribution in the synthetic data sample.
We see this situation in early work on using social media data to predict mental health conditions. Models to distinguish PTSD from depression turned out to mainly capture the differences in user age and gender, rather than language reflecting the actual conditions~\cite{preoctiuc2015role}.

\subsection{Other Bias Definitions and Frameworks}
While this is the first attempt at a comprehensive conceptual framework for bias in NLP, alternative frameworks exist, both in other fields and based on more qualitative definitions.
\newcite{friedler2016possibility} define bias as unfairness in algorithms. They specify the idea of a ``construct'' space, which captures the latent features in the data that help predict the right outcomes. 
They suggest that finding those latent variables would also enable us to produce the right outcomes.
\newcite{hovy2016social} take a broader scope on bias based on ethics in new technologies. They list three qualitative sources (data, modeling, and research design), and suggest three corresponding types of biases: demographic bias, overgeneralization, and topic exposure.
\newcite{suresh2019framework} propose a qualitative framework for bias in machine learning, defining bias as a ``potential harmful property of the data''. They categorize bias into historical bias, representation bias, measurement bias, and evaluation bias. 
\newcite{Glymour:2019:MBM:3287560.3287573} classify algorithmic bias, in general, into four different categories, depending on the causal conditional dependencies to which it is sensitive: procedural bias, outcome bias, behavior-relative error bias, and score-relative error bias. 
\newcite{corbett2018measure} propose statistical limitations of the three prominent definitions of fairness (anti-classification, classification parity, and calibration), enabling researchers to develop fairer machine learning algorithms.

Our framework focuses on NLP, but it follows \newcite{Glymour:2019:MBM:3287560.3287573} in providing probabilistic based definitions of bias. It incorporates and formalizes the above to varying degrees.

In social sciences, bias definitions often relate to the ability to test causal hypotheses. 
\newcite{hernan2004structural} propose a common structure for various types of selection bias. They define bias as the difference between a variable and the outcome, and the causal effect of a variable on the outcome. E.g., when the causal risk ratio (CRR) differs from associational risk ratio (ARR). 
Similarly, \newcite{baker2013summary} define bias as uncontrolled covariates or ``disturbing variables'' that are related to measures of interest.

Others provide definitions restricted to particular applications. For example, \newcite{caliskan2017semantics} propose the Word-Embedding Association Test (WEAT). It quantifies semantic bias based on the distance between words with demographic associations in the embedding space. The previously mentioned work by \newcite{kurita2019measuring} and \newcite{sweeney-najafian-2019-transparent} extend such measures. 
Similarly, \newcite{romanov2019s} define bias based on the correlation between the embeddings of human attributes with the difference in the True Positive rates between human traits. This approach is reflective of an error disparity. 

Our framework encompasses bias-related work in the social sciences. Please see the supplement in \ref{app:sec:relatedWork} for a brief overview.

\section{Countermeasures}
We group proposed countermeasures based on the origin(s) on which they act. 

\paragraph{Label Bias.}
There are several ways to address label bias, typically by controlling for biases of the annotators~\cite{pavlick2014language}. 
Disagreement between annotators has long been an active research area in NLP, with various approaches to measure and quantify disagreement through inter-annotator agreement (IAA) scores to remove outliers~\cite{artstein2008inter}. Lately, there has been more of an emphasis on embracing variation through the use of Bayesian annotation models~\cite{hovy2013learning,passonneau2014benefits,paun2018comparing}. These models arrive at a much less biased estimate for the final label than majority voting, by attaching confidence scores to each annotator, and reweighting them through that method.
Other approaches have explored harnessing the inherent disagreement among annotators to guide the training process~\cite{plank2014learning}. By weighting updates by the amount of disagreement on the labels, this method prevents bias towards any one label. The weighted updates act as a regularizer during training, which might also help prevent overamplification.
If annotators behave in predictable ways to produce artifacts (i.e., always add ``not'' to form a contradiction), we can train a model on such biased features and use it in ensemble learning \cite{clark2019don}.
\newcite{hays2015use} attempt to make Web studies equivalent to representative focus group panels. They give an overview of probabilistic and non-probabilistic approaches to generate the Internet panels that contribute to the data generation. Along with the six demographic attributes (age, gender, race/ethnicity, education, marital status, and income), they use poststratification to reduce the bias (some of these methods cross into addressing selection bias).

\paragraph{Selection bias.}
The primary source for selection bias is the mismatch between the sample distribution and the ideal distribution. Consequently, any countermeasures need to re-align the two distributions to minimize this mismatch.

The easiest way to address the mismatch is to re-stratify the data to more closely match the ideal distribution. However, this often involves downsampling an overly represented class, which reduces the number of available instances.
\newcite{mohammady2014using} use a stratified sampling technique to reduce the selection bias in the data. 
\newcite{almeida2015twitter} use demographic user attributes, including age, gender, and social status, to predict the election results in six different cities of Brazil. They use stratified sampling on all the resulting groups to reduce selection bias. 

Rather than re-sampling, others use \textit{reweighting} or \textit{poststratifying} to reduce selection bias.
\newcite{culotta2014reducing} estimates county-level health statistics based on social media data.
He shows we can stratify based on external socio-demographic data about a community's composition (e.g., gender and race).
\newcite{park2006state} estimate state-wise public opinions using the National Surveys corpus. To reduce bias, they use various socioeconomic and demographic attributes (state of residence, sex, ethnicity, age, and education level) in a multilevel logistic regression. 
\newcite{choy2011sentiment} and \newcite{choy2012us} also use race and gender as features for reweighting in predicting the results of the Singapore and US presidential elections. 
\newcite{baker2013summary} study how selection bias manifests in inferences for a larger population, and how to avoid it. Apart from the basic demographic attributes, they also consider attitudinal and behavioral attributes for the task.
They suggest using reweighting, ranking reweighting or propensity score adjustment, and sample-matching techniques to reduce selection bias.

Others have suggested combinations of these approaches. 
\newcite{hernan2004structural}, propose Directed Acyclic graphs for various heterogeneous types of selection bias, and suggest using stratified sampling, regression adjustment, or inverse probability weighting to avoid the bias in the data. 
\newcite{zagheni2015demographic}, study the use of Internet Data for demographic studies and propose two approaches to reduce the selection bias in their task. If the ground truth is available, they adjust selection bias based on the calibration of a stochastic microsimulation. If unavailable, they suggest using a difference-in-differences technique to find out trends on the Web.

\newcite{zmigrod-etal-2019-counterfactual} show that gender-based selection bias could be addressed by data augmentation, i.e., by adding slightly altered examples to the data.
This addition addresses selection bias originating in the features ($X_{source}$), so that the model is fit on a more gender-representative sample.
Their approach is similar to the reweighting of poll data based on demographics, which can be applied more directly to tweet-based population surveillance (see our last case study, \ref{appendix:case5}).

\newcite{li2018towards} introduce a model-based countermeasure. They use an adversarial multitask-learning setup to model demographic attributes as auxiliary tasks explicitly. By reversing the gradient for those tasks during backpropagation, they effectively force the model to ignore confounding signals associated with the demographic attributes. Apart from improving overall performance across demographics, they show that it also protects user privacy. The findings from \newcite{elazar2018adversarial}, however, suggest that even with adversarial training, internal representations still retain traces of demographic information.

\paragraph{Overamplification.}
In its simplest form, overamplification of inherent bias by the model can be corrected by downweighting the biased instances in the sample, to discourage the model from exaggerating the effects.

A common approach involves using synthetic matched distributions. 
To address gender bias in neural network approaches to coreference resolution~\newcite{rudinger2018gender, zhao2018gender} suggest matching the label distributions in the data, and training the model on the new data set. They swap male and female instances and merge them with the original data set for training. 
In the same vein, \newcite{webster2018mind} provide a gender-balanced training corpus for coreference resolution. Based on the first two corpora, \newcite{stanovsky-etal-2019-evaluating} introduce a bias evaluation for machine translation, showing that most systems overamplify gender bias (see also \newcite{prates2018assessing}). \newcite{hovy-etal-2020-can} show that this overamplification consistently makes translations sound older and more male than the original authors.

Several authors have suggested it is essential for language to be understood within the context of the author and their social environment~\newcite{jurgens2013s,danescu2013no,hovy2018social,yang2019seekers}.
Considering the author demographics improves the accuracy of text classifiers\newcite{volkova2013exploring,hovy2015demographic,lynn2017human}, and in turn, could lead to decreased error disparity.

\paragraph{Semantic bias.}
Countermeasures for semantic bias in embeddings typically attempt to adjust the parameters of the embedding model to reflect a target distribution more accurately. 
Because all of the above techniques can be applied for model fitting, here we highlight techniques that are more specific to addressing bias in embeddings. 

\newcite{bolukbasi2016man} suggest that techniques to de-bias embeddings can be classified into two approaches: hard de-biasing (completely removes bias) and soft de-biasing (partially removes bias avoiding side effects). 
\newcite{romanov2019s} generalize this work to a multi-class setting, exploring methods to mitigate bias in an occupation classification task. 
They reduce the correlation between the occupation of people and the word embedding of their names, and manage to simultaneously reduce race and gender biases without reducing the classifier's performance. 
\newcite{manzini2019black}, identify the bias subspace using principal component analysis and remove the biased components using hard Neutralize and Equalize de-biasing and soft biasing methods proposed by \newcite{bolukbasi2016man}. 
The above examples evaluate success through the semantic analogy task~\cite{mikolov2013efficient}, a method whose informativeness has since been questioned, though~\cite{nissim2019fair}. For a dedicated overview of semantic de-biasing techniques see \newcite{lauscher2019general}.

\paragraph{Social-Level Mitigation.}
Several initiatives propose standardized documentation to trace potential biases, and to ultimately mitigate them. \textit{Data Statements}~\newcite{bender2018data} suggest clearly disclosing data selection, annotation, and curation processes explicitly and transparently. 
Similarly,~\newcite{gebru2018datasheets} suggest \textit{Datasheets} to cover the lifecycle of data including ``motivation for dataset creation; dataset composition; data collection
process; data preprocessing; dataset distribution; dataset
maintenance; and legal and ethical considerations''. 
\newcite{mitchell2019model} extend this idea to include model specifications and performance details on different user groups.
\newcite{hitti-etal-2019-proposed} propose a taxonomy for assessing the gender bias of a data set.
While these steps do not directly mitigate bias, they can encourage researchers to identify and communicate sources of label or selection bias. 
Such documentation, combined with a conceptual framework to guide specific mitigation techniques, acts as an essential mitigation technique at the level of the research community.

See Appendix \ref{sec:caseStudies} for case studies outlining various types of bias in several NLP tasks.

\section{Conclusion}
We present a comprehensive overview of the recent literature on predictive bias in NLP. 
Based on this survey, we develop a unifying conceptual framework to describe bias sources and their effects (rather than \textit{just} their effects). This framework allows us to group and compare works on countermeasures.
Rather than giving the impression that bias is a growing problem, we would like to point out that bias is not necessarily something gone awry, but rather something nearly inevitable in statistical models. We do, however, stress that we need to acknowledge and address bias with proactive measures. Having a formal framework of the causes can help us achieve this.

We would like to leave the reader with these main points:
(1) every predictive model with errors is bound to have disparities over human attributes (even those not directly integrating human attributes);
(2) disparities can result from a variety of origins --- the embedding model, the feature sample, the fitting process, and the outcome sample --- within the standard predictive pipeline; 
(3) selection of ``protected attributes'' (or human attributes along which to avoid biases) is necessary for measuring bias, and often helpful for mitigating bias and increasing the generalization ability of the models.

We see this paper as a step toward a unified understanding of bias in NLP. We hope it inspires further work in both identifying and countering bias, as well as conceptually and mathematically defining bias in NLP. 

\section*{Framework Application Steps (TL;DR)}
\begin{flushleft}
\begin{enumerate}
\setlength{\itemsep}{0pt}%
\setlength{\parskip}{0pt}
\item Specify \textit{target} population and an \textit{ideal distribution} of the attribute ($A$) to be investigated for bias; 
Consult datasheets and data statements\footnote{~\cite{gebru2018datasheets,bender2018data}} if available for the model \textit{source};
\item If \textit{outcome disparity \textup{or} error disparity}, check for potential origins: 
\begin{enumerate}
 \setlength{\itemsep}{0pt}%
 \setlength{\parskip}{0pt}
 \item if \textit{label bias}: use post-stratification or retrain annotators.
 \item if \textit{selection bias}: use stratified sampling to match source to target populations, or use post-stratification, re-weighting techniques.  
 \item if \textit{overamplification}: synthetically match distributions or add outcome disparity to cost function.
 \item if \textit{semantic bias}: retrain or retrofit embeddings considering approaches above, but with attributed (e.g., gendered) words (rather than people) as the population. 
    \end{enumerate}
 \end{enumerate}
 \end{flushleft}

\section*{Acknowledegments}
The authors would like to thank Vinod Prabhakaran, Niranjan Balasubramanian, Joao Sedoc, Lyle Ungar, Rediet Abebe, Salvatore Giorgi, Margaret Kern and the anonymous reviewers for their constructive comments.
Dirk Hovy is a member of the Bocconi Institute for Data Science and Analytics (BIDSA) and the Data and Marketing Insights (DMI) unit.

\bibliographystyle{acl_natbib}
\bibliography{main} 
\appendix

\section{Appendices}
\label{sec:appendix}

\subsection{Related Work in Other Fields}
\label{app:sec:relatedWork}
We survey the literature of adjacent fields and outline different streams related to our framework. These examples illustrate the ubiquity and complexity of bias, and highlight its understanding in different disciplines over time.

Bias became a crucial topic in social science following the seminal work of Tversky and Kahneman, showing that human thinking was subject to systematic errors~\cite{tversky1973availability}. Human logic was seemingly separate from the principles of probability calculus. ``Bias'' here is interpreted as the result of psychological heuristics, i.e., mental ``shortcuts'' to help us react faster to situations. 
While many of these heuristics can be useful in critical situations, their indiscriminate application in everyday life can have adverse effects and cause bias. 
This line of work has spawned an entire field of study in psychology (decision making). 

The focus of~\newcite{tversky1973availability} (and a whole field of decision making that followed) was human behavior. Still, the same basic principles of systematic differences in decision making apply to machines as well. 
However, algorithms also provide systematic ways to reduce bias, and some see the mitigation of bias in algorithm decisions as a potential opportunity to move the needle positively ~\cite{kleinberg2018discrimination}.
Thus, we can apply frameworks of contemporaries in human behavior to machines~\cite{rahwan2019machine}, and perhaps benefit from a more scalable experimentation process. \newcite{costello2014surprisingly} studies human judgment under uncertain conditions, and proposes that we \textit{can} algorithmically account for observed human bias, provided there is sufficient random noise in the probabilistic model.
This view suggests bias within the model itself, what we have called \textit{overamplification}.

Still, most works on bias in decision making assume working with unbiased data, even though social science has long battled selection bias. 
Most commonly, data selection is heavily skewed towards the students found on western university campuses~\cite{henrich2010weirdest}. 
Attempts to remedy selection bias in a scalable fashion use online populations, which in turn are skewed by unequal access to the Internet, but which we can mitigate through reweighting schemes ~\cite{couper2013sky}.

In some cases, algorithmic bias has helped understand society better. 
For example, \textit{semantic bias} in word embeddings has been leveraged to track trends in societal attitudes concerning gender roles and ethnic stereotypes. ~\newcite{garg2018word,kozlowski2018geometry} measure the distance between certain sets of words in different decades to track this change.
This use of biased embeddings illustrates an interesting distinction between \textit{normative} and \textit{descriptive} ethics. 
When used in predictive models, semantic bias is something to be avoided \cite{bolukbasi2016man}. I.e., it is \textit{normatively wrong} for many applications (e.g., we ideally would want all genders or ethnicities equally associated with all jobs). 
However, the works by \newcite{garg2018word} and \newcite{kozlowski2018geometry} show that it is precisely this bias of word embeddings that reflects societal attitudes. Here, the presence of bias is \textit{descriptively correct}. Similarly, \newcite{bhatia2017associative} uses this property of word embeddings to measure people's psychological biases and attitudes towards making individual decisions.

\subsection{Discussion: Example Case Studies}
\label{sec:caseStudies}

\paragraph*{Part of Speech Taggers and Parsing.}
\label{appendix:case1}
The works by \newcite{hovy2015tagging, jorgensen2015challenges} outline the effect of selection bias on syntactic tools. The language of demographic groups systematically differs from each other for syntactic attributes. Therefore, models trained on samples whose demographic composition (e.g., age and ethnicity) differs from the target perform significantly worse.
Within the predictive bias framework, the consequence of this selection bias is an \textit{error disparity} -- $Q(\epsilon_{D = general}|A = {age, ethnicity}) \nsim Uniform$, the error of the model across a general domain ($D$) is not uniform with respect to attributes ($A$) age and ethnicity. 
\newcite{li2018towards} shows that this consequence of selection bias can be addressed by adversarial learning, removing the age gap and significantly reducing the performance difference between ethnolects (even if it was not trained with that objective).
\newcite{garimella-etal-2019-womens} quantifies this bias further by studying the effect of different gender compositions of the training data on tagging and parsing, supporting the claim that debiased samples benefit performance.

\paragraph*{Image Captions.}
\label{appendix:case2}
\newcite{hendricks2018women} shows the presence of gender bias in image captioning, overamplifying differences present in the training data. Prior work focused on context (e.g., it is easier to predict ``mouse'' when there is a computer present). This bias manifests in ignoring people present in the image. The gender bias is not only influenced by the images, but also by biased language models. The primary consequence is an \textit{outcome disparity} -- $Q(\hat{Y}_D|gender) \nsim P(Y_D|gender)$, the distribution of outcomes (i.e. caption words and phrases) produced from the model $Q(\hat{Y}_{D}|gender)$ over-selects particular phrases beyond the distribution observed in reality: (i.e.  $P(Y_{D}|gender)$; this is true even when the source and target are the same: $D = source = target$). 

To overcome the bias and to increase performance, \newcite{hendricks2018women} introduce an equalizer model with two loss-terms: Appearance Confusion Loss (ACL) and Confident Loss (Conf). ACL increases the gender confusion when gender information is not present in the image, making it difficult to predict an accurately gendered word. Confident loss increases the confidence of the predicted gendered word when gender information \textit{is} present in the image. Both loss terms have the effect of decreasing the difference between $Q(\hat{Y}_D|gender)$ and $P(\hat{Y}_D|gender)$. In the end, the Equalizer model performs better in predicting a woman while still misclassifying a man as a woman, but decreasing \textit{error disparity} overall. 

\paragraph*{Sentiment Analysis.}
\label{appendix:case3}
\newcite{kiritchenko2018examining} show the issues of both semantic bias and overamplification. They assess scoring differences in 219 sentiment analysis systems by switching out names and pronouns. (They switch between male and female pronouns, and between prototypical white and black American first names based on name registers.) The results show that male pronouns are associated with higher scores for negative polarity, and prototypical black names with higher scores for negative emotions. The consequence of the semantic bias and overamplification are outcome disparities:  $Q(\hat{Y}_D|gender) \nsim P(Y_D|gender)$ and $Q(\hat{Y}_D|race) \nsim P(Y_D|race)$. 
This finding again demonstrates a case of descriptive vs.\ normative ethics. We could argue that because aggression is more often associated with male protagonists, the models reflect a descriptively correct (if morally objectionable) societal fact. However, if the model score changes based on ethnicity, the difference likely reflects (and amplifies) societal ethnic stereotypes. Those stereotypes, though, are both normatively and descriptively wrong. 

\paragraph*{Differential Diagnosis in Mental Health.}
\label{appendix:case4}
In the clinical community, differentiating a subject with post-traumatic stress disorder (PTSD) from someone with depression is known to be difficult. It was, therefore, surprising when early work on this task produced AUCs greater than 0.85 (this and similar tasks were part of the CLPsych2015 Shared task; \cite{coppersmith2015clpsych}).
Labels of depression and PTSD had been automatically derived from a convenience sample of individuals\footnote{A convenience sample, a term from social science, is a set of data selected because it is available rather than designed for the given task.} who had publicly stated their diagnosis in their profile. The task included a 50/50 split from each category.  
However, \newcite{preotiuc2015mental} show that these classifiers primarily picked up on differences in age or gender -- subjects with PTSD were more likely to be older than those with depression. 
While age and gender themselves are valid information for mental health diagnosis, the design yielded classifiers that predicted nearly all older individuals to have PTSD, and those younger to have depression, despite the 50/50 split. These classifiers resulted in \textit{outcome disparity}, because older individuals were much less likely to be labeled depressed than the target population (and younger less likely for PTSD:  $Q(\hat{Y}_{D}|A = age) \nsim P(Y_{D}|A = age)$).
In the end, the task organizers mitigated the issue by using matched controls -- adding another 50\% samples for each class such that the age and gender distributions of both groups matched. 
Recently, \newcite{benton2017multitask} showed that accounting for demographic attributes in the model could leverage this correlation while controlling for the confounds.

\paragraph*{Assessing Demographic Variance in Language.}
\label{appendix:case5}
A final case study in applying our framework demonstrates how inferring user demographics can mitigate bias. 
Consider the task of producing population measurements from readily available (but biased) community corpora. E.g., assessing representative US county life satisfaction from tweets~\cite{schwartz2013characterizing}. 
Unlike our other examples, the \textit{outcomes} of the source training data (i.e., surveys) are expected to be representative, while the \textit{features} come with biases. The source feature distributions with respect to human attributes are dissimilar from the ideal distribution, while the source outcomes match that target outcomes ($Q(X_{source}|A) \nsim P(X_{target}|A)$ but $Q(Y_{source}|A) \sim P(Y_{target}|A)$). 

In this case, the effectiveness of countermeasures preventing selection and semantic biases (for $X_{source}$ and $X_{target}$) should result in increased predictive performance against a representative community outcome. 
Indeed, \newcite{giorgi2019correcting} adjust the feature estimates, $X$, to match representative demographics and socio-economics by using inferred user attributes, and find improved predictions for the life satisfaction of a Twitter community. 



\end{document}